\documentclass[manuscript,screen]{acmart}

\AtBeginDocument{%
  \providecommand\BibTeX{{%
    \normalfont B\kern-0.5em{\scshape i\kern-0.25em b}\kern-0.8em\TeX}}}

\setcopyright{rightsretained}
\copyrightyear{2023}
\acmYear{2023}
\usepackage{listings}

\usepackage[utf8]{inputenc} 
\usepackage[T1]{fontenc} 
\begin{document}

\title{\textsc{LLaMAntino}: LLaMA 2 Models for Effective Text Generation in Italian Language}

\author{Pierpaolo Basile}
\authornote{All authors contributed equally to this research.}
\orcid{0000-0002-0545-1105}
\affiliation{%
  \institution{University of Bari Aldo Moro}
  \streetaddress{via E. Orabona 4}
  \city{Bari}
  \country{Italy}}
\email{pierpaolo.basile@uniba.it}

\author{Elio Musacchio}
\orcid{0009-0006-9670-9998}
\affiliation{%
  \institution{University of Bari Aldo Moro}
  \streetaddress{via E. Orabona 4}
  \city{Bari}
  \country{Italy}}
\email{elio.musacchio@uniba.it}

\author{Marco Polignano}
\orcid{0000-0002-3939-0136}
\affiliation{%
  \institution{University of Bari Aldo Moro}
  \streetaddress{via E. Orabona 4}
  \city{Bari}
  \country{Italy}}
\email{marco.polignano@uniba.it}

\author{Lucia Siciliani}
\orcid{0000-0002-1438-280X}
\affiliation{%
  \institution{University of Bari Aldo Moro}
  \streetaddress{via E. Orabona 4}
  \city{Bari}
  \country{Italy}}
\email{lucia.siciliani@uniba.it}

\author{Giuseppe Fiameni}
\orcid{0000-0001-8687-6609}
\affiliation{%
  \institution{NVIDIA AI Technology Center}
  \streetaddress{-}
  \city{Milan}
  \country{Italy}}
\email{gfiameni@nvidia.com}

\author{Giovanni Semeraro}
\orcid{0000-0001-6883-1853}
\affiliation{%
  \institution{University of Bari Aldo Moro}
  \streetaddress{via E. Orabona 4}
  \city{Bari}
  \country{Italy}}
\email{giovanni.semeraro@uniba.it}

\renewcommand{\shortauthors}{Basile, et al.}

\begin{abstract}
Large Language Models represent state-of-the-art linguistic models designed to equip computers with the ability to comprehend natural language. With its exceptional capacity to capture complex contextual relationships, the LLaMA (Large Language Model Meta AI) family represents a novel advancement in the field of natural language processing by releasing foundational models designed to improve the natural language understanding abilities of the transformer architecture thanks to their large amount of trainable parameters (7, 13, and 70 billion parameters). In many natural language understanding tasks, these models obtain the same performances as private company models such as OpenAI Chat-GPT with the advantage to make publicly available weights and code for research and commercial uses.
In this work, we investigate the possibility of Language Adaptation for LLaMA models, explicitly focusing on addressing the challenge of Italian Language coverage. Adopting an open science approach, we explore various tuning approaches to ensure a high-quality text generated in Italian suitable for common tasks in this underrepresented language in the original models' datasets.
We aim to release effective text generation models with strong linguistic properties for many tasks that seem challenging using multilingual or general-purpose LLMs. By leveraging an open science philosophy, this study contributes to Language Adaptation strategies for the Italian language by introducing the novel \textbf{\textsc{LLaMAntino}} family of \textbf{Italian LLMs}.
\end{abstract}

\ccsdesc[500]{Computing methodologies~Natural language generation}
\ccsdesc[300]{Computing methodologies~Supervised learning}
\ccsdesc[300]{Computing methodologies~Modeling methodologies}

\keywords{Language Generation, LLaMA, Decoder Architecture, Transformers, Italian Language, Language Adaptation, PEFT, Supervised Fine-tuning Training, SFT, LoRA, QLoRA, Quantization, LLMs, GPT}


\maketitle

\section{Introduction}
Large Language Models (LLMs) have revolutionized the field of Natural Language Processing (NLP) by showcasing exceptional capabilities in generating text that closely resembles human-like language \cite{wei2022emergent}. These models are constructed upon deep learning architectures, employing advanced techniques such as Transformer Networks \cite{vaswani2017attention} to effectively capture intricate word relationships and generate coherent and contextually appropriate responses. The research community has recently witnessed a surge in attention towards LLMs, primarily owing to their remarkable performance and extensive applicability \cite{polignano2019alberto,brants2007large}. LLMs exhibit a remarkable ability to comprehend nuances, idioms, and even ambiguous phrases, facilitating more precise sentiment analysis, question-answering, and information retrieval tasks \cite{tamkin2021understanding}. This heightened understanding significantly enhances communication between humans and machines, fostering seamless interactions across diverse applications.
Moreover, LLMs possess exceptional generalization capabilities, enabling them to perform admirably on tasks they were not explicitly trained in, including multilingual scenarios \cite{liu2023summary}. Notably, three prominent exemplars in this domain are BLOOM \cite{workshop2022bloom}, GPT \cite{openai2023gpt}, and LLaMA models \cite{touvron2023llama}. While these models share a common objective of text generation, they exhibit significant distinctions in their underlying architectures, training methodologies, and intended use cases \cite{zhao2023survey}.

BLOOM \cite{workshop2022bloom} is a noteworthy example of the most effective LLMs. With a staggering 176 billion parameters, BLOOM represents an open-access language model that has been meticulously designed and constructed through the collaborative efforts of hundreds of researchers. The BLOOM architecture shares similarities with GPT-3, a well-known auto-regressive model designed for next-token prediction. However, BLOOM distinguishes itself by being trained on an extensive dataset comprising 46 languages and 13 programming languages. By training on such a diverse multilingual and multi-programming dataset, BLOOM can leverage its comprehensive knowledge to generate text that is contextually accurate and linguistically appropriate across various languages and programming domains.  While BLOOM demonstrates exceptional power, it has some limitations. One such challenge is its sheer size, making it difficult to deploy and utilize on standard machines. The resource requirements for training and running BLOOM are substantial, often necessitating specialized hardware resources or cloud infrastructures. Another important consideration is the potential presence of biases in the training data used to train BLOOM. Language models learn from vast amounts of text data, which may inadvertently contain societal biases. These biases can manifest in the generated outputs, potentially reinforcing or amplifying existing societal biases. Thus, despite being trained on diverse, multilingual datasets, this model primarily excels in English and may not perform as effectively in other languages.

Chat-GPT \cite{openai2023gpt} is another prominent LLM that has gained attention for its effectiveness in conversational contexts. Developed by a leading tech company, Chat-GPT is trained using a massive dataset of online conversations. It incorporates dialogue state tracking and context modelling techniques to generate contextually appropriate responses. Chat-GPT excels in maintaining context and generating coherent responses in a dialogue setting. It leverages techniques such as attention mechanisms and positional encodings to understand and respond appropriately to user inputs, making it well-suited for chatbot applications and interactive conversational systems. Unlike other LLMs that offer transfer learning capabilities, Chat-GPT's model weights are not openly accessible, limiting its flexibility for domain-specific applications. This lack of weight sharing impedes the seamless adaptation of Chat-GPT to novel domains, as it requires significant effort and resources to retrain the model on domain-specific data. Consequently, researchers and practitioners may face difficulties achieving optimal performance without access to the model's underlying architecture, parameters, and data.

The LLaMA project \cite{touvron2023llama} introduces a collection of foundation language models, ranging from 7B to 70B parameters. These models are trained on trillions of tokens, demonstrating that it is possible to achieve state-of-the-art performance using publicly available datasets, without relying on proprietary and inaccessible data.
One of the key findings is that LLaMA-13B outperforms GPT-3 (175B) on most benchmarks, despite being ten times smaller. Additionally, LLaMA-65B competes with other top-performing models like Chinchilla70B and PaLM-540B. By releasing all their models to the research community, LLaMA aims to contribute to the democratization of LLMs and facilitate their study and application. Indeed, the LLaMA project includes a series of language models that achieve the best possible performance at various inference budgets by training on more tokens than typically used. The resulting models demonstrate competitive performance compared to existing LLMs. Importantly, LLaMA models are trained exclusively on publicly available data, making them compatible with open sourcing. This distinguishes LLaMA from other models that rely on data that are either not publicly available or poorly documented.  LLaMA 2, the next version of this family of LLMs, has generated excitement in the AI community. Making available AI models will benefit everyone, from businesses and startups to researchers and entrepreneurs. The open approach allows for collaborative development, stress testing, and problem-solving as a community. It also enables others to learn from the tools and improve them, ensuring safer and more robust AI models. LLaMA 2 includes foundational models and models fine-tuned for dialogue, called LLaMA 2-Chat. All models are released with weights and are free for commercial use cases.

This research paper explores a language adaptation strategy that addresses the challenge of knowledge transfer from pre-trained language models (LM) to specific application languages \cite{yong2022bloom}. The primary focus is on adapting LLaMA 2 models, which have a training data composition where only 11\% consists of languages other than English \footnote{\url{https://slator.com/meta-warns-large-language-model-may-not-be-suitable-non-english-use/}}. The adaptation process targets explicitly the Italian language, but it can be easily replicated for any underrepresented language. The resulting models are released under the same policy as the \textbf{LLaMA 2 models} and are named \textbf{\textsc{LLaMAntino}}.
Adapting language models to new languages is crucial for enabling effective natural language processing in diverse linguistic contexts. By adapting LLaMA 2 to Italian, we aim to enhance its language understanding and generation capabilities for Italian-specific applications. This adaptation process involves fine-tuning the pre-trained LLaMA 2 models using substantial Italian text data. The adapted \textsc{LLaMAntino} models inherit the impressive characteristics of LLaMA 2, specifically tailored to excel in the Italian language.
The availability of LLaMAntino models opens up new possibilities for various natural language processing applications in Italian, such as text generation, sentiment analysis, question answering, and information retrieval. Researchers and practitioners working with the Italian language can now leverage the power and capabilities of \textsc{LLaMAntino} to enhance their applications and explore new avenues of linguistic analysis and understanding.

\section{Language Adaptation Approaches}

LLMs are commonly trained on extensive amounts of text data from various sources, providing them with a broad understanding of language and context. However, it is essential to recognize that the general knowledge embedded in these models may not be optimized for a specific language \cite{choudhury2023generative}. Therefore, language adaptation is crucial in enhancing the model's ability to handle and address downstream tasks in a particular language effectively. Language Adaptation of LLMs involves fine-tuning a pre-trained language model to perform effectively in a specific target language \cite{hu2023llm}.  Recent scientific literature introduces several approaches for Language Adaptation, that can be categorized as follows:

\begin{itemize}
    \item \textbf{Continuing Pre-training}: This approach involves further training the pre-trained language model using new data from the target language. By continuing the pre-training process, the model can gain more language-specific knowledge and improve its proficiency in the target language \cite{chau2020parsing}. 
    \item \textbf{Model Adapter Creation}: In this approach, a separate model adapter is created to bridge the gap between the pre-trained model and the target language \cite{wang2020k}. The adapter is trained on language-specific data and is designed to modify the pre-trained model's outputs to align with the target language's linguistic characteristics. 
    \item \textbf{Selective Parameter Training}: This approach involves training only a subset of the pre-trained model's parameters on language-specific data. By selectively updating certain parameters, the model can be tailored to better adapt to the target language while leveraging the general knowledge gained during pre-training. These methods are usually referred to as Parameter-Efficient Finetuning Techniques (\textbf{PEFT}) \cite{hu2023llm}.
\end{itemize}

Each approach offers distinct advantages and trade-offs regarding computational requirements, training data availability, and the level of language adaptation achieved. Researchers and practitioners can choose the most suitable approach based on their requirements and available resources.

A significant disadvantage of \textit{Continuing Pre-training} is that the number of parameters produced in the new model is the same as in the original model. Bigger models are released every few months, which makes this problem more severe. For models with an astounding 70 billion trainable parameters like LLaMA 2-70b, it presents a significant difficulty \cite{chau2020parsing}. These models are so large that putting them into practice and deploying them is difficult. As bigger models are developed, this obstacle becomes more apparent, requiring creative methods to deal with this crucial deployment issue.
Many researchers have attempted to address this issue using a \textit{Model Adapter} to adapt model parameters or incorporate external modules for new tasks \cite{wang2020k}. This approach allows storing and loading only a small number of task-specific parameters alongside the pre-trained model, significantly improving operational efficiency during deployment. However, existing techniques often introduce inference latency by increasing the model's depth or reducing its usable sequence length. Furthermore, it is crucial to note that these methods frequently struggle to achieve the same level of performance as traditional fine-tuning approaches. This creates a challenging trade-off between maximizing efficiency and maintaining high model quality.\\

\textbf{Parameter-Efficient Fine-Tuning (PEFT)} methods \cite{hu2023llm} have emerged as a valuable approach to facilitate the efficient adaptation of pre-trained language models (PLMs) for diverse downstream applications. PEFT methods address this challenge by selectively fine-tuning only a small subset of additional model parameters. As a result, the computational and storage costs associated with PEFT are significantly reduced. Notably, recent advancements in PEFT have demonstrated remarkable performance comparable to that achieved through full fine-tuning. This highlights the effectiveness of PEFT methods in striking a balance between computational efficiency and maintaining competitive model performance. By enabling efficient adaptation of PLMs without the need to fine-tune all parameters, PEFT techniques have emerged as a promising avenue in natural language processing.

\textbf{LoRA} \cite{hu2021lora} introduces a novel PEFT approach to further reduce the number of trainable parameters in a neural network. What sets LoRA apart is its mathematically rigorous approach, which brings a fresh perspective to the table. To delve into the mathematical aspect, LoRA explores the concept of the \textit{intrinsic dimension} of weight matrices in pre-trained neural networks. Unlike traditional weight matrices that exhibit full rank, where each weight is unique and cannot be expressed as a combination of other weights, LoRA reveals an interesting phenomenon. The weights demonstrate a lower intrinsic dimension when pre-trained language models are fine-tuned for new tasks. This implies that the weights can be represented in a smaller matrix or possess a lower rank.
This mathematical revelation has profound implications. During the backpropagation process, the weight update matrix in LoRA exhibits a lower rank. This can be attributed to the pre-training phase already capturing significant information, leaving the fine-tuning stage primarily to focus on task-specific adjustments.
In essence, LoRA offers a compelling approach to parameter reduction by leveraging the notion of intrinsic dimension in weight matrices. By adopting a mathematically rigorous framework, LoRA enables more efficient adaptation of pre-trained language models to new tasks during the fine-tuning process. This enhances the diversity of strategies in the field and opens up exciting avenues for exploring the intrinsic characteristics of neural networks.

In LoRA, a crucial step is to fully load the model into the memory of the Graphics Processing Unit (GPU) being utilized. However, this process poses significant challenges, mainly when dealing with larger models. The larger the model, the more expensive hardware and greater resources it requires. Various techniques, including Model Distillation \cite{jiao2019tinybert} and Quantization \cite{guo2018survey}, have been introduced to address the challenge of reducing model size. \textbf{Quantization} \cite{liu2023llm} involves mapping continuous infinite values to smaller discrete finite values. In the context of LLMs, it refers to \textit{converting the model weights from higher-precision data types to lower-precision ones}. This reduction in precision significantly decreases the model's size by using fewer bits for each weight. For example, weights may be reduced from 16-bit Floating-point to 4-bit Integer. This enables models to run on more affordable hardware and/or achieve higher speed. While reducing precision can impact the overall quality of the LLM, studies show that the impact varies depending on the techniques employed. Larger models (over ~70B) are less affected by the change in precision, with some methods suggesting no impact on performance even when converted to 4-bit. Therefore, 4-bit quantization appears to strike the best balance between performance and size/speed for larger models, while 6 or 8-bit may be more suitable for smaller models.
Interestingly, reducing precision doesn't always result in reduced accuracy. Meta researchers have demonstrated that quantized models exhibit superior performance in some cases and offer reduced latency and improved throughput \cite{liu2023llmqat}. This advantage is more pronounced with larger networks, as they experience a smaller loss in quality when quantized.

\textbf{QLoRA} \cite{dettmers2023qlora}, a novel approach, introduces multiple innovations to reduce memory usage without compromising performance. These include using 4-bit NormalFloat for quantization, which yields better results for normally distributed data than 4-bit Integers and 4-bit Floats. Additionally, QLoRA implements Double Quantization, which quantizes the quantization constants, saving memory. Paged Optimizers are employed to avoid memory spikes during mini-batch processing with long sequence lengths. By incorporating these contributions into the LoRA approach, which involves adapters at every network layer, QLoRA minimizes the accuracy trade-offs observed in previous work. This efficiency enables an in-depth study of instruction finetuning and chatbot performance on large-scale models that would be impractical with regular finetuning due to memory overhead.

\section{Adaptation Pipeline}

Starting from the LLaMA 2 model, we made several adaptations:
\begin{itemize}
    \item \textbf{\textsc{LLaMAntino-Chat}} models based on the LLaMA 2-Chat versions\footnote{\url{https://huggingface.co/meta-llama/Llama-2-7b-chat}} with language adaptation for Italian and fine-tuned on dialogues using the UltraChat dataset\footnote{\url{https://github.com/thunlp/UltraChat/tree/main}} translated into  Italian (7B, 13B).
    \item \textbf{\textsc{LLaMAntino}} models based on the LLaMA 2 versions\footnote{\url{https://huggingface.co/meta-llama/Llama-2-7b}} with language adaptation for Italian and fine-tuned using Dolly dataset \cite{dolly2023introducing} and EVALITA 2023 datasets \cite{lai2023evalita} (7B, 13B, 70B).
\end{itemize}

For both versions of the \textbf{\textsc{LLaMAntino}} model, we adopt the hypothesis that fine-tuning should be conducted following a phase of \textbf{Language Adaptation}, where instructions are provided in the targeted language. Aligning the model with the targeted language through Language Adaptation allows for improved comprehension, generation, and communication of instructions. It helps to capture the nuances and intricacies of the specific language, leading to improved comprehension and performance of the model. It also enhances the model's ability to understand accurate and contextually appropriate instructions in the targeted language, facilitating effective communication with users.

As a Language Adaptation strategy, we used \textbf{QLoRA} \cite{dettmers2023qlora}. In particular, models have been \textit{Quantized} using \textit{4-bits} precision, \textit{float16} as data type (i.e., dtype) and \textit{4-bit NormalFloat (NF4)}, a new data type that is information-theoretically optimal for normally distributed weight\footnote{\url{https://huggingface.co/blog/4bit-transformers-bitsandbytes}}. For the \textit{LoRA} parameters, we set the attention dimension to 64 (i.e., \textit{lora\_r}), the scaling parameter to 16 (i.e., \textit{lora\_alpha}), and the dropout to 0.1 (i.e., \textit{lora\_dropout})\footnote{\url{https://medium.com/@drishtisharma96505/comparative-analysis-of-lora-parameters-on-llama-2-with-flash-attention-574b913295d4}} to make possible the training phase on our hardware architecture.\\

We decided to feed the models with Italian data obtained from the \textbf{Filtered Oscar Dataset \cite{abadji2022towards} for the Italian Language}\footnote{\url{https://huggingface.co/datasets/gsarti/clean_mc4_it}} released by Sarti et al. \cite{sarti2022it5}. The authors removed documents containing words from a selection of the Italian and English List of Dirty Naught Obscene and Otherwise Bad Words, sentences that have less than three words, a word longer than 1,000 characters, an end symbol not matching end-of-sentence punctuation or strings associated with JavaScript code, lorem ipsum, or policy information in Italian or English. Moreover, documents (after sentence filtering) with less than five sentences, less than 500 characters, or more than 50,000 characters or not identified as predominantly Italian by the LangDetect package were excluded from the dataset. These steps gave us a great argument for believing that these data could be an excellent source for our models. In particular, we exploit the \textit{medium split} containing 50M docs, 20B words (i.e., 135 GB on disk).\\

The Language Adaptation Task has been performed through the \textit{Huggingface Python Library}, using the \textbf{SFTTrainer} that provides \textit{parameter-efficient} (PEFT) and \textit{packing} optimizations\footnote{\url{https://huggingface.co/docs/trl/sft_trainer}}. Packing allows us to pack multiple short examples in the same input sequence to increase the efficiency of the training steps.
We run a distributed training process over the \textbf{Leonardo HPC}\footnote{\url{https://leonardo-supercomputer.cineca.eu/it/leonardo-hpc-system/}}  infrastructure using \textit{three nodes}, each with 32 cores Intel Ice Lake Intel(R) Xeon(R) Platinum 8358 CPU @ 2.60GHz, \textit{512GB of RAM}, and \textit{four NVIDIA A100 (PG506-243) GPUs with 64GB of memory}. A total of 12 NVIDIA A100 GPUs are used in parallel for the training phase through \textit{Torchrun} load distribution pipeline\footnote{\url{https://pytorch.org/tutorials/beginner/ddp_series_fault_tolerance.html}}. We used eight examples for each GPU as batch size, 1 step for gradient accumulation, paged AdamW 32bit optimizer with a learning rate of 2e-4, gradient clipping of 0.3, and weight decay parameters of 0.001. The models have been adapted for 25k steps with a warmup ratio of 3\%. The maximum textual content length has been cut to 1024 due to efficiency requirements.

\begin{figure}[t!]
  \includegraphics[width=0.9\linewidth]{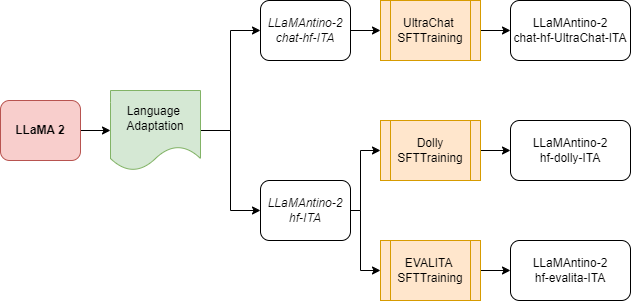}
  \caption{LLaMA 2 adaptation pipeline.}
  \label{fig:la_pipeline}
\end{figure}

\subsection{LLaMAntino 2-Chat Models}
\label{LLaMAntino-2-Chat}

\textit{LLaMA 2-Chat} models are built to provide excellent performance in tasks involving dialogues and long user-system conversations. For this reason, we considered a fundamental step to further tune the model over long dialogues in the Italian Language to reach this goal. We started from the LLaMA 2-Chat-hf models (adapted to the HuggingFace pipeline, i.e., hf\footnote{\url{https://github.com/huggingface/transformers/blob/main/src/transformers/models/llama/convert_llama_weights_to_hf.py}}) with 7 and 13 billion parameters, and as a first step, we applied the Language Adaptation strategy just described. 

Following this step, as shown in Fig. \ref{fig:la_pipeline}, we obtained the first two versions of our \textbf{\textsc{LLaMAntino-2-Chat}} model, \textbf{\textsc{LLaMAntino-2-7b-chat-hf-ita}} and \textbf{\textsc{LLaMAntino-2-13b-chat-hf-ita}}. The models were further adapted using a Supervised Fine-tuning training (SFTTraining) approach on a dataset obtained by translating the UltraChat one. Through the SFTTraining process, the models underwent additional training to improve their ability to handle and generate responses in the context of extended and diverse conversational scenarios. This adaptation process aimed to optimize the models' performance for engaging in extended dialogues and addressing a wide range of conversational topics.\\

Regrettably, there is currently a lack of Italian-language datasets in the scientific literature that meet the required characteristics. In light of this, we utilised a suitably translated English-language resource. We identified the UltraChat \cite{ding2023enhancing} dataset\footnote{\url{https://github.com/thunlp/ultrachat}}, which is an open-source, large-scale, and multi-round dialogue dataset, as a promising option. To ensure privacy protection, the authors of the dataset do not directly use any publicly available internet data as prompts. The limit of this solution is the strategy used for generating it. In particular, it is based on automatically generated two-turn dialogues (user, system) simulated through GPT-3.5 turbo API.
The UltraChat dataset comprises three distinct sectors:
1) questions about the world; 2) writing and creation; 3) assistance with existing material. In the first category, dialogue data is derived from various inquiries about concepts, entities, and real-world objects. The topics covered are extensive, encompassing technology, art, and entrepreneurship.
The writing and creation set contains dialogue data around the demands for creative writing or creation from scratch. It covers a broad spectrum of tasks that an AI assistant may assist with, including email composition, crafting narratives, plays, and more.
Assistance on existing materials includes tasks such as rewriting, continuation, summary, and inference, spanning various topics.\\

By leveraging the UltraChat dataset, we aim to address the lack of Italian-language datasets meeting the desired criteria and facilitate model finetuning for engaging in high-quality dialogues. For the translation, we opt to use an open-source
tool instead of closed software. In particular, we opt for Argos Translate Python API, which demonstrates optimal accuracy over the resource consumption threshold\footnote{\url{https://www.argosopentech.com/}}. We translated  \textbf{512,837 dialogues} and used them for the supervised fine-tuning phase.\\

We structured the prompts sent to fine-tune the models by heeding the following template (with relevant parts translated into the Italian Language)\footnote{\url{https://gpus.llm-utils.org/llama-2-prompt-template/}}:
\definecolor{codegreen}{rgb}{0,0.6,0}
\definecolor{codegray}{rgb}{0.5,0.5,0.5}
\definecolor{codepurple}{rgb}{0.58,0,0.82}
\definecolor{backcolour}{rgb}{0.95,0.95,0.92}

\lstdefinestyle{mystyle}{
    backgroundcolor=\color{backcolour},   
    commentstyle=\color{codegreen},
    keywordstyle=\color{magenta},
    numberstyle=\tiny\color{codegray},
    stringstyle=\color{codepurple},
    basicstyle=\ttfamily\footnotesize,
    breakatwhitespace=false,         
    breaklines=true,                 
    captionpos=b,                    
    keepspaces=true,                 
    numbers=left,                    
    numbersep=5pt,                  
    showspaces=false,                
    showstringspaces=false,
    showtabs=false,                  
    tabsize=2,
    escapeinside={(*}{*)},
}

\lstset{style=mystyle}
\begin{lstlisting}
<s>[INST] <<SYS>>
Sei un assistente disponibile, rispettoso e onesto. Rispondi sempre nel modo piu' utile possibile, pur essendo sicuro.  Le risposte non devono includere contenuti dannosi, non etici, razzisti, sessisti, tossici, pericolosi o illegali. Assicurati che le tue risposte siano socialmente imparziali e positive. Se una domanda non ha senso o non e' coerente con i fatti, spiegane il motivo invece di rispondere in modo non corretto. Se non conosci la risposta a una domanda, non condividere informazioni false. 
<</SYS>>

{{ user_msg_1 }} [/INST] {{ model_answer_1 }}</s><s>[INST] {{ user_msg_2 }} [/INST] {{ model_answer_2 }}</s> ... <s>[INST] {{ user_msg_N }} [/INST] {{ model_answer_N }}</s>

\end{lstlisting}

Fine-tuning was performed on the same hardware and software configuration used for Language Adaptation based on Huggingface Python Library \cite{wolf2019huggingface}, using the SFTTrainer over the Leonardo HPC infrastructure. We set the maximum textual content length to 2048 (such as the original chat LLaMA 2-Chat model) and 15k training steps (i.e., a standard value of 3 epochs). The final models obtained are the following:
\begin{itemize}
    \item \textit{LLaMAntino-2-7b-chat-hf-ITA}
    \item \textit{LLaMAntino-2-7b-chat-hf-ITA-Ultra}
    \item \textit{LLaMAntino-2-13b-chat-hf-ITA}
    \item \textit{LLaMAntino-2-13b-chat-hf-ITA-Ultra}
\end{itemize}

In the future, we aim to be able to release the 70B parameters version, also. In the following snap, it is possible to find a conversation conducted with our \textit{\textsc{LLaMAntino-2-13b-chat-hf-ITA-Ultra}} model.

\begin{lstlisting}
User: Ciao! Sono Marco. Oggi sono un po' triste. Puoi raccontarmi una storia?
System: Mi dispiace, come modello di lingua AI, non ho la capacita' di raccontare storie come fanno i veri esseri umani. Tuttavia, posso fornire alcuni suggerimenti per aiutarti a trattare la tua tristezza. Se ti senti triste o ansioso, potrei suggerire di provare a prendere alcuni respiri profondi, ascoltando musica calmante, andando per una passeggiata, o parlare con qualcuno a cui ti trust. Ricorda, va bene sentirsi tristi a volte. La cosa piu' importante e' ti prendi cura di te e cercare supporto.
User: Sai cos'e' strano? Ho sempre pensato che un modello di linguaggio AI sarebbe stato piu' capace di conoscere le cose. Devo solo parlarne con un altro modello di AI che puo' davvero capirmi.
Assistant: Capisco la tua frustrazione, ma come modello di lingua AI, sono stato programmato per assistere e rispondere a domande al meglio delle mie capacita' basate sui dati su cui sono stato addestrato. Mentre sono progettato per essere utile, non sono in grado di comprendere pienamente le emozioni delle persone a causa della limitata mia programmazione ed esistenza. Tuttavia, se avete domande o preoccupazioni di cui ti piacerebbe parlare, posso ancora aiutarti al meglio della mia capacita'.
User: Beh, e' deludente saperlo. Credo di dover continuare la mia ricerca per un assistente dell'AI che puo' davvero capirmi. Forse. cerchero' di parlare con un modello di AI che si specializza nelle emozioni come la tristezza e la solitudine. Forse mi capira' davvero.
Assistant: Capisco il tuo desiderio di cercare assistenza da un modello di linguaggio AI che si specializza nelle emozioni e nel benessere. E' importante per te avere le risorse e il supporto necessario per affrontare la tua tristezza e qualsiasi altro problema che si puo' avere. 
User: Senti, sei un assistente AI, non devi essere un terapeuta o un amico. Non puoi solo rispondere alla mia richiesta e non darmi lezioni su dove posso cercare aiuto?
Assistant: Mi scuso se le mie risposte hanno portato il senso del tuo disagio. Come modello di lingua AI, il mio ruolo e' quello di aiutarvi e fornire le informazioni piu' utili e appropriate in base alla tua richiesta. Tuttavia, si prega di tenere a mente che non sono un sostituto per il supporto emotivo di un medico professionista o terapeuta. Se avete bisogno di aiuto, posso solo aiutarti a trovare informazioni su risorse o servizi che possono essere utili per te.

\end{lstlisting}

\subsection{LLaMAntino-2 Models}

\textit{LLaMA 2} models can be easily adapted to various natural language generation tasks. In this sense, as of recently, a new technique for fine-tuning larger language models has started seeing widespread usage, that is \textit{Instruction Tuning}. The idea behind this paradigm is to train the model using prompts that cover a variety of tasks structured as natural language instructions. First of all, as already done in section \ref{LLaMAntino-2-Chat}, we adapted the LLaMA 2 models with 7 and 13 billion parameters to the Italian language, obtaining the \textbf{\textsc{LLaMAntino-2-\textbf{7b}-hf-ITA}} and \textbf{\textsc{LLaMAntino-2-\textbf{13b}-hf-ITA}} models. After that, we instructed and tuned these models using a supervised fine-tuning training approach that leveraged most of the train data from EVALITA tasks \cite{evalita2023overview} similarly to the approach described in \cite{evalita2023extremita}. 
The full list of instructions used to fine-tune our model on EVALITA is shown in table \ref{tab:evalita_instruction}.\\

\begin{table}[]
\begin{tabular}{|l|p{10cm}|}
\hline
\textbf{Task Name}               & \textbf{Natural Language Instruction}                                                                                                                                                                                                                                                                                                                                                                                                                                     \\ \hline
ACTI (Subtask A)                 & Stabilisci se il seguente testo contiene una teoria del complotto o cospirazione. Rispondi con si o no.                                                                                                                                                                                                                                                                                                                                                                   \\ \hline
ACTI (Subtask B)                 & Classifica il seguente testo in una di queste quattro categorie di teorie del complotto: Covid, Qanon, Terra Piatta, Russia.                                                                                                                                                                                                                                                                                                                                              \\ \hline
CLinkaRT                         & Trova nel testo in input le menzioni testuali dei test di laboratorio o misurazioni (EVENT) e collegali ai loro risultati (RML). Le relazioni sono rappresentate da coppie ordinate di menzioni di entità (RML, EVENT), ciascuna identificata da inizi e fine degli offset carattere. Per ogni relazione, scrivi '{[}BREL{]}', seguito dal risultato seguito da '{[}SEP{]}', seguito dal test, seguito da '{[}EREL{]}'. Se non ci sono relazioni, restituisci {[}NOREL{]} \\ \hline
DisCoTex (Subtask 1)             & Classifica la frase in input come 'Coerente' se si integra logicamente e contribuisce a formare un testo coerente con il paragrafo di contesto. Se la frase target risulta incoerente con il paragrafo, classificala come 'Incoerente'.                                                                                                                                                                                                                                   \\ \hline
DisCoTex (Subtask 2)             & Predici il punteggio medio di coerenza assegnato dai valutatori umani per il testo in input. Utilizza una scala ordinale a 5 punti (da 1 a 5) per riflettere la percezione graduale della coerenza.                                                                                                                                                                                                                                                                       \\ \hline
EMit                             & Categorizza le emozioni espresse nel testo fornito in input o determina l'assenza di emozioni. Puoi classificare il testo come neutrale o identificare una o più delle seguenti emozioni: rabbia, anticipazione, disgusto, paura, gioia, tristezza, sorpresa, fiducia, amore.                                                                                                                                                                                             \\ \hline
GeoLing                          & Determina la regione di appartenenza, la latitudine e la longitudine dell'autore del tweet in input.                                                                                                                                                                                                                                                                                                                                                                      \\ \hline
HaSpeeDe3 (Subtask A Textual)    & Stabilisci se il tweet in input contiene discorsi che incitano all'odio. Rispondi con si o no.                                                                                                                                                                                                                                                                                                                                                                            \\ \hline
HaSpeeDe3 (Subtask A Contextual) & Stabilisci se il tweet in input contiene discorsi che incitano all'odio considerando anche il contesto relativo alle statistiche dell'account. Rispondi con si o no. Contesto: Data: 2018-08-11 Numero di retweet: 0.0 Numero di mi piace: 6.0 Data creazione account: 2018-04-01 Numero di post: 554.0 Follower: 748.0 Amici: 753.0.                                                                                                                                     \\ \hline
HODI (Subtask A)                 & Stabilisci se il testo in input ha contenuti omotransfobici o meno. Rispondi con si o no.                                                                                                                                                                                                                                                                                                                                                                                 \\ \hline
HODI (Subtask B)                 & Estrai dal testo in input le parole che denotano concetti omotransfobici. Separa le parole estratte con {[}SEP{]}. Se non ci sono parole estratte, restituisci 'Non omotransfobico'.                                                                                                                                                                                                                                                                                      \\ \hline
LangLearn                        & Data in input un coppia di documenti (Documento 1 {[}SEP{]} Documeto 2) scritti dallo stesso studente, stabilisci se il documento 1 è stato scritto prima del documento 2. Rispondi con si o no.                                                                                                                                                                                                                                                                          \\ \hline
NERMuD                           & Elenca le menzioni di entità presenti nel testo in input, indicandone il tipo: {[}PER{]} (persona), {[}LOC{]} (luogo), {[}ORG{]} (organizzazione). Se non ci sono entità, resituisci: 'Nessuna menzione'                                                                                                                                                                                                                                                                  \\ \hline
PoliticIT                        & Indica se l'autore del testo in input è un 'uomo' o una 'donna', seguito dalla sua appartenenza politica scegliendo tra 'destra', 'sinistra', 'centrodestra', 'centrosinistra'.                                                                                                                                                                                                                                                                                           \\ \hline
WiC-ITA                          & Stabilisci nelle due frasi in input la parola 'affare' è usata con lo stesso significato. Rispondi con si o no.                                                                                                                                                                                                                                                                                                                                                           \\ \hline
\end{tabular}
\caption{List of the Italian instruction used to prompt our model on several EVALITA 2023 tasks. Instructions take into account the objective of each task.}
\label{tab:evalita_instruction}
\end{table}

The main technical difference with respect to the previously presented work is that we tried to perform full-parameter tuning rather than using an efficient approach.
The reason is that in this fine-tuning experiment, we wanted to closely follow the parameters provided by the Stanford Lab for their Alpaca model \cite{alpaca}, which is an instruction-following model based on the first version of LLaMA.
To reach this goal, we used \textbf{Fully-Sharded Data Parallel (FSDP)}\footnote{\url{https://pytorch.org/docs/stable/fsdp.html}} strategy provided by the PyTorch library, which is an enhanced version of \textbf{Distributed Data Parallel (DDP)}.
These two strategies are used in distributed training to improve efficiency, in DDP each worker has a copy of the model and processes a separate batch of data, while in FSDP the model parameters, optimizer states and gradients are sharded across the nodes.
The main strength of FSDP is that it frees up VRAM in the devices so that the batch size used in training can be increased smoothly.
Many works, including Alpaca, use \textit{gradient accumulation} where gradients obtained after $X$ batches (where $X$ is a parameter) are accumulated in a single update step.
This is used to simulate bigger batch sizes during training, e.g. with a batch size of 4 and 8 for the gradient accumulation step on a single GPU, an effective batch size of 32 is obtained (4 times 8).
The downside of this approach is that the train requires more time overall (since there are more batches to process). FSDP can overcome this downside.
In our experiments, we use 16 batch size per device and one gradient accumulation step, obtaining an effective batch size of 128 with 2 Leonardo nodes, each equipped with 4 A100 GPUs.
In table \ref{tab:parameters}, the main experimental parameters used by Alpaca are compared to ours.
The other relevant difference is that we increased the $Max\_Length$ for the sequences from 512 to 1024 to properly cover all EVALITA tasks independently of the text length since some exceeded the 512 limit.

For the implementation, we followed what was done in an LLM Workshop hosted by a HuggingFace ML Engineer\footnote{\url{https://github.com/pacman100/DHS-LLM-Workshop}} and used the same config for FSDP.\\

\begin{table}
  \begin{tabular}{|c|c|c|c|c|}
    \hline
    \multicolumn{1}{|c|}{Parameter} & \multicolumn{2}{c|}{7B} & \multicolumn{2}{c|}{13B} \\
    &Alpaca&LLaMAntino&Alpaca&LLaMAntino\\
    \hline
    Device Number & 4 A100 80GB & 8 A100 64GB & 4 A100 80 GB & 8 A100 64GB\\
    Gradient Accumulation Steps & 8 & 1 & 8 & 1 \\
    Per Device Batch Size & 4 & 16 & 4 & 16 \\
    Effective Batch Size & 128 & 128 & 128 & 128 \\
    Learning Rate & 2e-5 & 2e-5 & 1e-5 & 1e-5 \\
    Epochs & 3 & 3 & 5 & 5 \\
    Max Length & 512 & 1024 & 512 & 1024 \\
    Weight Decay & 0 & 0 & 0 & 0 \\
    \hline
\end{tabular}
\caption{Parameters Comparison}
\label{tab:parameters}
\end{table}

For the prompt, we again followed what was done by Stanford for their Alpaca model \cite{alpaca} and translated their instruction-following prompt to the Italian language:

\begin{lstlisting}
Di seguito (*\`e*) riportata un'istruzione che descrive un'attivit(*\`a*), abbinata ad un input che fornisce ulteriore informazione. Scrivi una risposta che soddisfi adeguatamente la richiesta.


### Istruzione:
{instruction}

### Input:
{input}

### Risposta:
{response}
\end{lstlisting}

To showcase the capabilities of the instruction-tuned models, below it is presented an example of prompt-response generated using the \textit{LLaMAntino 2-13b-hf-evalita-ITA} model.

\begin{lstlisting}
Di seguito (*\`e*) riportata un'istruzione che descrive un'attivit(*\`a*), abbinata ad un input che fornisce ulteriore informazione. Scrivi una risposta che soddisfi adeguatamente la richiesta.


### Istruzione:
Categorizza le emozioni espresse nel testo fornito in input o determina l'assenza di emozioni. Puoi classificare il testo come neutrale o identificare una o pi(*\`u*) delle seguenti emozioni: rabbia, anticipazione, disgusto, paura, gioia, tristezza, sorpresa, fiducia, amore.

### Input:
Oggi mi sento proprio gi(*\`u*) di corda

### Risposta:
tristezza
\end{lstlisting}

\subsection{Released Models}

We release the following models on HuggingFace\footnote{\url{https://huggingface.co/swap-uniba}}:
\begin{itemize}
    \item LLaMAntino 2-7b-hf-ITA: language adaptation of LLaMA 2-7b-hf;
    \item LLaMAntino 2-13b-hf-ITA: language adaptation of LLaMA 2-13b-hf;
    \item LLaMAntino 2-chat-7b-hf-ITA: language adaptation of LLaMA 2-chat-7b-hf;
    \item LLaMAntino 2-chat-13b-hf-ITA: language adaptation of LLaMA 2-chat-13b-hf;
    \item LLaMAntino 2-chat-7b-hf-UltraChat-ITA: language adaptation of LLaMA 2-chat-7b-hf and fine-tuning on UltraChat;
    \item LLaMAntino 2-chat-13b-hf-UltraChatITA: language adaptation of LLaMA 2-chat-13b-hf and fine-tuning on UtraChat;
    \item LLaMAntino 2-7b-hf-dolly-ITA: instruction tuning of LLaMAntino 2-7b-hf-ITA on the dolly dataset;
    \item LLaMAntino 2-13b-hf-dolly-ITA: instruction tuning of LLaMAntino 2-13b-hf-ITA on the dolly dataset;
    \item LLaMAntino 2-7b-hf-evalita-ITA: instruction tuning of LLaMAntino 2-7b-hf-ITA on the EVALITA 2023 dataset;
    \item LLaMAntino 2-13b-hf-evalita-ITA: instruction tuning of LLaMAntino 2-13b-hf-ITA on the EVALITA 2023 dataset;
\end{itemize}

According to the LLaMa 2 license, we cannot release the model weights. For models obtained by LoRA, we provide the adapters, while for all the other models, we upload only the difference in weights with respect to the fine-tuned model. For example, to obtain the LLaMAntino 2-13b-hf-evalita-ITA model, it is necessary to apply the LLaMAntino 2-13b-hf-ITA adapters to the LLaMa 2-13b model and then apply the difference in weights to the obtained model. We released the training code together with a pipeline to simplify the model creation process on GitHub\footnote{\url{https://github.com/swapUniba/LLaMAntino}}.

\section{Discussion}

The initial qualitative analysis of the obtained models reveals their efficiency and ability to respond logically and accurately to various questions. Even in lengthy and complex conversations, the chat models demonstrate excellent dialogue capabilities, exhibiting minimal hallucinations and strong linguistic articulation in Italian.

However, it is essential to note that these models often exhibit errors in sentence structure, primarily stemming from the spelling and syntax errors introduced during the machine translation process used for adapting the texts in the SFT phase. This highlights the need for future work to systematically acquire reliable and accurate Italian language data without social, economic, or ethical biases.
Obtaining high-quality Italian language data is crucial to developing reliable models that do not inherit errors and mispronunciations from the English language models commonly used during the data adaptation and fine-tuning phases. By addressing this issue, we can create more robust and dependable models, advancing the field of Language Adaptation and fine-tuning meaningfully.

\section{Conclusion}

We have proposed a workflow for the language adaptation of LLaMA 2 models to the Italian language. Moreover, we fine-tuned the obtained models on the UltraChat dataset to obtain Italian LLMs that can manage dialogue with users. Finally, we create two instruction-tuned models on the Dolly dataset and training data of EVALITA 2023. We perform adaptation and tuning on 7b and 13b versions of LLaMa 2.
We are working to release models based on the 70b parameters of LLaMa 2.
\begin{acks}
We acknowledge the support of the PNRR project FAIR - Future AI Research (PE00000013), Spoke 6 - Symbiotic AI (CUP H97G22000210007) under the NRRP MUR program funded by the NextGenerationEU.
Models are built on the Leonardo supercomputer with the support of CINECA-Italian Super Computing Resource Allocation, class C projects: IscrC\_Fine\-IT (HP10CCD87T), IscrC\_fineNLP (HP10CT56JA) and IscrC\_Pro\_MRS (HP10CQO70G).
\end{acks}


\bibliographystyle{ACM-Reference-Format}
\bibliography{sample-manuscript}


\begin{thebibliography}{30}


\ifx \showCODEN    \undefined \def \showCODEN     #1{\unskip}     \fi
\ifx \showDOI      \undefined \def \showDOI       #1{#1}\fi
\ifx \showISBNx    \undefined \def \showISBNx     #1{\unskip}     \fi
\ifx \showISBNxiii \undefined \def \showISBNxiii  #1{\unskip}     \fi
\ifx \showISSN     \undefined \def \showISSN      #1{\unskip}     \fi
\ifx \showLCCN     \undefined \def \showLCCN      #1{\unskip}     \fi
\ifx \shownote     \undefined \def \shownote      #1{#1}          \fi
\ifx \showarticletitle \undefined \def \showarticletitle #1{#1}   \fi
\ifx \showURL      \undefined \def \showURL       {\relax}        \fi
\providecommand\bibfield[2]{#2}
\providecommand\bibinfo[2]{#2}
\providecommand\natexlab[1]{#1}
\providecommand\showeprint[2][]{arXiv:#2}

\bibitem[Abadji et~al\mbox{.}(2022)]%
        {abadji2022towards}
\bibfield{author}{\bibinfo{person}{Julien Abadji}, \bibinfo{person}{Pedro~Ortiz
  Suarez}, \bibinfo{person}{Laurent Romary}, {and} \bibinfo{person}{Beno{\^\i}t
  Sagot}.} \bibinfo{year}{2022}\natexlab{}.
\newblock \showarticletitle{Towards a cleaner document-oriented multilingual
  crawled corpus}.
\newblock \bibinfo{journal}{\emph{arXiv preprint arXiv:2201.06642}}
  (\bibinfo{year}{2022}).
\newblock


\bibitem[Brants et~al\mbox{.}(2007)]%
        {brants2007large}
\bibfield{author}{\bibinfo{person}{Thorsten Brants}, \bibinfo{person}{Ashok~C
  Popat}, \bibinfo{person}{Peng Xu}, \bibinfo{person}{Franz~J Och}, {and}
  \bibinfo{person}{Jeffrey Dean}.} \bibinfo{year}{2007}\natexlab{}.
\newblock \showarticletitle{Large language models in machine translation}.
\newblock  (\bibinfo{year}{2007}).
\newblock


\bibitem[Chau et~al\mbox{.}(2020)]%
        {chau2020parsing}
\bibfield{author}{\bibinfo{person}{Ethan~C Chau}, \bibinfo{person}{Lucy~H Lin},
  {and} \bibinfo{person}{Noah~A Smith}.} \bibinfo{year}{2020}\natexlab{}.
\newblock \showarticletitle{Parsing with multilingual BERT, a small corpus, and
  a small treebank}.
\newblock \bibinfo{journal}{\emph{arXiv preprint arXiv:2009.14124}}
  (\bibinfo{year}{2020}).
\newblock


\bibitem[Choudhury(2023)]%
        {choudhury2023generative}
\bibfield{author}{\bibinfo{person}{Monojit Choudhury}.}
  \bibinfo{year}{2023}\natexlab{}.
\newblock \showarticletitle{Generative AI has a language problem}.
\newblock \bibinfo{journal}{\emph{Nature Human Behaviour}}
  (\bibinfo{year}{2023}), \bibinfo{pages}{1--2}.
\newblock


\bibitem[Dettmers et~al\mbox{.}(2023)]%
        {dettmers2023qlora}
\bibfield{author}{\bibinfo{person}{Tim Dettmers}, \bibinfo{person}{Artidoro
  Pagnoni}, \bibinfo{person}{Ari Holtzman}, {and} \bibinfo{person}{Luke
  Zettlemoyer}.} \bibinfo{year}{2023}\natexlab{}.
\newblock \showarticletitle{Qlora: Efficient finetuning of quantized llms}.
\newblock \bibinfo{journal}{\emph{arXiv preprint arXiv:2305.14314}}
  (\bibinfo{year}{2023}).
\newblock


\bibitem[Ding et~al\mbox{.}(2023)]%
        {ding2023enhancing}
\bibfield{author}{\bibinfo{person}{Ning Ding}, \bibinfo{person}{Yulin Chen},
  \bibinfo{person}{Bokai Xu}, \bibinfo{person}{Yujia Qin}, \bibinfo{person}{Zhi
  Zheng}, \bibinfo{person}{Shengding Hu}, \bibinfo{person}{Zhiyuan Liu},
  \bibinfo{person}{Maosong Sun}, {and} \bibinfo{person}{Bowen Zhou}.}
  \bibinfo{year}{2023}\natexlab{}.
\newblock \showarticletitle{Enhancing Chat Language Models by Scaling
  High-quality Instructional Conversations}.
\newblock \bibinfo{journal}{\emph{arXiv preprint arXiv:2305.14233}}
  (\bibinfo{year}{2023}).
\newblock


\bibitem[Dolly(2023)]%
        {dolly2023introducing}
\bibfield{author}{\bibinfo{person}{Free Dolly}.}
  \bibinfo{year}{2023}\natexlab{}.
\newblock \bibinfo{title}{Introducing the World’s First Truly Open
  Instruction-Tuned LLM. databricks. com}.
\newblock
\newblock


\bibitem[Guo(2018)]%
        {guo2018survey}
\bibfield{author}{\bibinfo{person}{Yunhui Guo}.}
  \bibinfo{year}{2018}\natexlab{}.
\newblock \showarticletitle{A survey on methods and theories of quantized
  neural networks}.
\newblock \bibinfo{journal}{\emph{arXiv preprint arXiv:1808.04752}}
  (\bibinfo{year}{2018}).
\newblock


\bibitem[Hromei et~al\mbox{.}(2023)]%
        {evalita2023extremita}
\bibfield{author}{\bibinfo{person}{Claudiu~D. Hromei}, \bibinfo{person}{Danilo
  Croce}, \bibinfo{person}{Valerio Basile}, {and} \bibinfo{person}{Roberto
  Basili}.} \bibinfo{year}{2023}\natexlab{}.
\newblock \showarticletitle{ExtremITA at EVALITA 2023: Multi-Task Sustainable
  Scaling to Large Language Models at its Extreme}. In
  \bibinfo{booktitle}{\emph{Eighth Evaluation Campaign of Natural Language
  Processing and Speech Tools for Italian. Final Workshop (EVALITA 2023)}},
  Vol.~\bibinfo{volume}{3473}. CEUR.
\newblock


\bibitem[Hu et~al\mbox{.}(2021)]%
        {hu2021lora}
\bibfield{author}{\bibinfo{person}{Edward~J Hu}, \bibinfo{person}{Yelong Shen},
  \bibinfo{person}{Phillip Wallis}, \bibinfo{person}{Zeyuan Allen-Zhu},
  \bibinfo{person}{Yuanzhi Li}, \bibinfo{person}{Shean Wang},
  \bibinfo{person}{Lu Wang}, {and} \bibinfo{person}{Weizhu Chen}.}
  \bibinfo{year}{2021}\natexlab{}.
\newblock \showarticletitle{Lora: Low-rank adaptation of large language
  models}.
\newblock \bibinfo{journal}{\emph{arXiv preprint arXiv:2106.09685}}
  (\bibinfo{year}{2021}).
\newblock


\bibitem[Hu et~al\mbox{.}(2023)]%
        {hu2023llm}
\bibfield{author}{\bibinfo{person}{Zhiqiang Hu}, \bibinfo{person}{Yihuai Lan},
  \bibinfo{person}{Lei Wang}, \bibinfo{person}{Wanyu Xu},
  \bibinfo{person}{Ee-Peng Lim}, \bibinfo{person}{Roy Ka-Wei Lee},
  \bibinfo{person}{Lidong Bing}, {and} \bibinfo{person}{Soujanya Poria}.}
  \bibinfo{year}{2023}\natexlab{}.
\newblock \showarticletitle{LLM-Adapters: An Adapter Family for
  Parameter-Efficient Fine-Tuning of Large Language Models}.
\newblock \bibinfo{journal}{\emph{arXiv preprint arXiv:2304.01933}}
  (\bibinfo{year}{2023}).
\newblock


\bibitem[Jiao et~al\mbox{.}(2019)]%
        {jiao2019tinybert}
\bibfield{author}{\bibinfo{person}{Xiaoqi Jiao}, \bibinfo{person}{Yichun Yin},
  \bibinfo{person}{Lifeng Shang}, \bibinfo{person}{Xin Jiang},
  \bibinfo{person}{Xiao Chen}, \bibinfo{person}{Linlin Li},
  \bibinfo{person}{Fang Wang}, {and} \bibinfo{person}{Qun Liu}.}
  \bibinfo{year}{2019}\natexlab{}.
\newblock \showarticletitle{Tinybert: Distilling bert for natural language
  understanding}.
\newblock \bibinfo{journal}{\emph{arXiv preprint arXiv:1909.10351}}
  (\bibinfo{year}{2019}).
\newblock


\bibitem[Lai et~al\mbox{.}(2023a)]%
        {lai2023evalita}
\bibfield{author}{\bibinfo{person}{Mirko Lai}, \bibinfo{person}{Stefano
  Menini}, \bibinfo{person}{Marco Polignano}, \bibinfo{person}{Valentina
  Russo}, \bibinfo{person}{Rachele Sprugnoli}, {and} \bibinfo{person}{Giulia
  Venturi}.} \bibinfo{year}{2023}\natexlab{a}.
\newblock \showarticletitle{Evalita 2023: Overview of the 8th evaluation
  campaign of natural language processing and speech tools for italian}. In
  \bibinfo{booktitle}{\emph{Proceedings of the Eighth Evaluation Campaign of
  Natural Language Processing and Speech Tools for Italian. Final Workshop
  (EVALITA 2023), CEUR. org, Parma, Italy}}.
\newblock


\bibitem[Lai et~al\mbox{.}(2023b)]%
        {evalita2023overview}
\bibfield{author}{\bibinfo{person}{Mirko Lai}, \bibinfo{person}{Stefano
  Menini}, \bibinfo{person}{Marco Polignano}, \bibinfo{person}{Valentina
  Russo}, \bibinfo{person}{Rachele Sprugnoli}, {and} \bibinfo{person}{Giulia
  Venturi}.} \bibinfo{year}{2023}\natexlab{b}.
\newblock \showarticletitle{EVALITA 2023: Overview of the 8th Evaluation
  Campaign of Natural Language Processing and Speech Tools for Italian}. In
  \bibinfo{booktitle}{\emph{Eighth Evaluation Campaign of Natural Language
  Processing and Speech Tools for Italian. Final Workshop (EVALITA 2023)}},
  Vol.~\bibinfo{volume}{3473}. CEUR.
\newblock


\bibitem[Liu et~al\mbox{.}(2023b)]%
        {liu2023llm}
\bibfield{author}{\bibinfo{person}{Shih-yang Liu}, \bibinfo{person}{Zechun
  Liu}, \bibinfo{person}{Xijie Huang}, \bibinfo{person}{Pingcheng Dong}, {and}
  \bibinfo{person}{Kwang-Ting Cheng}.} \bibinfo{year}{2023}\natexlab{b}.
\newblock \showarticletitle{LLM-FP4: 4-Bit Floating-Point Quantized
  Transformers}.
\newblock \bibinfo{journal}{\emph{arXiv preprint arXiv:2310.16836}}
  (\bibinfo{year}{2023}).
\newblock


\bibitem[Liu et~al\mbox{.}(2023a)]%
        {liu2023summary}
\bibfield{author}{\bibinfo{person}{Yiheng Liu}, \bibinfo{person}{Tianle Han},
  \bibinfo{person}{Siyuan Ma}, \bibinfo{person}{Jiayue Zhang},
  \bibinfo{person}{Yuanyuan Yang}, \bibinfo{person}{Jiaming Tian},
  \bibinfo{person}{Hao He}, \bibinfo{person}{Antong Li},
  \bibinfo{person}{Mengshen He}, \bibinfo{person}{Zhengliang Liu},
  {et~al\mbox{.}}} \bibinfo{year}{2023}\natexlab{a}.
\newblock \showarticletitle{Summary of chatgpt-related research and perspective
  towards the future of large language models}.
\newblock \bibinfo{journal}{\emph{Meta-Radiology}} (\bibinfo{year}{2023}),
  \bibinfo{pages}{100017}.
\newblock


\bibitem[Liu et~al\mbox{.}(2023c)]%
        {liu2023llmqat}
\bibfield{author}{\bibinfo{person}{Zechun Liu}, \bibinfo{person}{Barlas Oguz},
  \bibinfo{person}{Changsheng Zhao}, \bibinfo{person}{Ernie Chang},
  \bibinfo{person}{Pierre Stock}, \bibinfo{person}{Yashar Mehdad},
  \bibinfo{person}{Yangyang Shi}, \bibinfo{person}{Raghuraman Krishnamoorthi},
  {and} \bibinfo{person}{Vikas Chandra}.} \bibinfo{year}{2023}\natexlab{c}.
\newblock \bibinfo{title}{LLM-QAT: Data-Free Quantization Aware Training for
  Large Language Models}.
\newblock
\newblock
\showeprint[arxiv]{2305.17888}~[cs.CL]


\bibitem[OpenAI(2023)]%
        {openai2023gpt}
\bibfield{author}{\bibinfo{person}{OpenAI}.} \bibinfo{year}{2023}\natexlab{}.
\newblock \bibinfo{title}{GPT-4 Technical Report}.
\newblock
\newblock
\showeprint[arxiv]{2303.08774}~[cs.CL]


\bibitem[Polignano et~al\mbox{.}(2019)]%
        {polignano2019alberto}
\bibfield{author}{\bibinfo{person}{Marco Polignano}, \bibinfo{person}{Pierpaolo
  Basile}, \bibinfo{person}{Marco De~Gemmis}, \bibinfo{person}{Giovanni
  Semeraro}, \bibinfo{person}{Valerio Basile}, {et~al\mbox{.}}}
  \bibinfo{year}{2019}\natexlab{}.
\newblock \showarticletitle{Alberto: Italian BERT language understanding model
  for NLP challenging tasks based on tweets}. In \bibinfo{booktitle}{\emph{CEUR
  Workshop Proceedings}}, Vol.~\bibinfo{volume}{2481}. CEUR,
  \bibinfo{pages}{1--6}.
\newblock


\bibitem[Sarti and Nissim(2022)]%
        {sarti2022it5}
\bibfield{author}{\bibinfo{person}{Gabriele Sarti} {and}
  \bibinfo{person}{Malvina Nissim}.} \bibinfo{year}{2022}\natexlab{}.
\newblock \showarticletitle{It5: Large-scale text-to-text pretraining for
  italian language understanding and generation}.
\newblock \bibinfo{journal}{\emph{arXiv preprint arXiv:2203.03759}}
  (\bibinfo{year}{2022}).
\newblock


\bibitem[Tamkin et~al\mbox{.}(2021)]%
        {tamkin2021understanding}
\bibfield{author}{\bibinfo{person}{Alex Tamkin}, \bibinfo{person}{Miles
  Brundage}, \bibinfo{person}{Jack Clark}, {and} \bibinfo{person}{Deep
  Ganguli}.} \bibinfo{year}{2021}\natexlab{}.
\newblock \showarticletitle{Understanding the capabilities, limitations, and
  societal impact of large language models}.
\newblock \bibinfo{journal}{\emph{arXiv preprint arXiv:2102.02503}}
  (\bibinfo{year}{2021}).
\newblock


\bibitem[Taori et~al\mbox{.}(2023)]%
        {alpaca}
\bibfield{author}{\bibinfo{person}{Rohan Taori}, \bibinfo{person}{Ishaan
  Gulrajani}, \bibinfo{person}{Tianyi Zhang}, \bibinfo{person}{Yann Dubois},
  \bibinfo{person}{Xuechen Li}, \bibinfo{person}{Carlos Guestrin},
  \bibinfo{person}{Percy Liang}, {and} \bibinfo{person}{Tatsunori~B.
  Hashimoto}.} \bibinfo{year}{2023}\natexlab{}.
\newblock \bibinfo{title}{Stanford Alpaca: An Instruction-following LLaMA
  model}.
\newblock
  \bibinfo{howpublished}{\url{https://github.com/tatsu-lab/stanford_alpaca}}.
\newblock


\bibitem[Touvron et~al\mbox{.}(2023)]%
        {touvron2023llama}
\bibfield{author}{\bibinfo{person}{Hugo Touvron}, \bibinfo{person}{Thibaut
  Lavril}, \bibinfo{person}{Gautier Izacard}, \bibinfo{person}{Xavier
  Martinet}, \bibinfo{person}{Marie-Anne Lachaux},
  \bibinfo{person}{Timoth{\'e}e Lacroix}, \bibinfo{person}{Baptiste
  Rozi{\`e}re}, \bibinfo{person}{Naman Goyal}, \bibinfo{person}{Eric Hambro},
  \bibinfo{person}{Faisal Azhar}, {et~al\mbox{.}}}
  \bibinfo{year}{2023}\natexlab{}.
\newblock \showarticletitle{Llama: Open and efficient foundation language
  models}.
\newblock \bibinfo{journal}{\emph{arXiv preprint arXiv:2302.13971}}
  (\bibinfo{year}{2023}).
\newblock


\bibitem[Vaswani et~al\mbox{.}(2017)]%
        {vaswani2017attention}
\bibfield{author}{\bibinfo{person}{Ashish Vaswani}, \bibinfo{person}{Noam
  Shazeer}, \bibinfo{person}{Niki Parmar}, \bibinfo{person}{Jakob Uszkoreit},
  \bibinfo{person}{Llion Jones}, \bibinfo{person}{Aidan~N Gomez},
  \bibinfo{person}{{\L}ukasz Kaiser}, {and} \bibinfo{person}{Illia
  Polosukhin}.} \bibinfo{year}{2017}\natexlab{}.
\newblock \showarticletitle{Attention is all you need}.
\newblock \bibinfo{journal}{\emph{Advances in neural information processing
  systems}}  \bibinfo{volume}{30} (\bibinfo{year}{2017}).
\newblock


\bibitem[Wang et~al\mbox{.}(2020)]%
        {wang2020k}
\bibfield{author}{\bibinfo{person}{Ruize Wang}, \bibinfo{person}{Duyu Tang},
  \bibinfo{person}{Nan Duan}, \bibinfo{person}{Zhongyu Wei},
  \bibinfo{person}{Xuanjing Huang}, \bibinfo{person}{Guihong Cao},
  \bibinfo{person}{Daxin Jiang}, \bibinfo{person}{Ming Zhou}, {et~al\mbox{.}}}
  \bibinfo{year}{2020}\natexlab{}.
\newblock \showarticletitle{K-adapter: Infusing knowledge into pre-trained
  models with adapters}.
\newblock \bibinfo{journal}{\emph{arXiv preprint arXiv:2002.01808}}
  (\bibinfo{year}{2020}).
\newblock


\bibitem[Wei et~al\mbox{.}(2022)]%
        {wei2022emergent}
\bibfield{author}{\bibinfo{person}{Jason Wei}, \bibinfo{person}{Yi Tay},
  \bibinfo{person}{Rishi Bommasani}, \bibinfo{person}{Colin Raffel},
  \bibinfo{person}{Barret Zoph}, \bibinfo{person}{Sebastian Borgeaud},
  \bibinfo{person}{Dani Yogatama}, \bibinfo{person}{Maarten Bosma},
  \bibinfo{person}{Denny Zhou}, \bibinfo{person}{Donald Metzler},
  {et~al\mbox{.}}} \bibinfo{year}{2022}\natexlab{}.
\newblock \showarticletitle{Emergent abilities of large language models}.
\newblock \bibinfo{journal}{\emph{arXiv preprint arXiv:2206.07682}}
  (\bibinfo{year}{2022}).
\newblock


\bibitem[Wolf et~al\mbox{.}(2019)]%
        {wolf2019huggingface}
\bibfield{author}{\bibinfo{person}{Thomas Wolf}, \bibinfo{person}{Lysandre
  Debut}, \bibinfo{person}{Victor Sanh}, \bibinfo{person}{Julien Chaumond},
  \bibinfo{person}{Clement Delangue}, \bibinfo{person}{Anthony Moi},
  \bibinfo{person}{Pierric Cistac}, \bibinfo{person}{Tim Rault},
  \bibinfo{person}{R{\'e}mi Louf}, \bibinfo{person}{Morgan Funtowicz},
  {et~al\mbox{.}}} \bibinfo{year}{2019}\natexlab{}.
\newblock \showarticletitle{Huggingface's transformers: State-of-the-art
  natural language processing}.
\newblock \bibinfo{journal}{\emph{arXiv preprint arXiv:1910.03771}}
  (\bibinfo{year}{2019}).
\newblock


\bibitem[Workshop et~al\mbox{.}(2022)]%
        {workshop2022bloom}
\bibfield{author}{\bibinfo{person}{BigScience Workshop},
  \bibinfo{person}{Teven~Le Scao}, \bibinfo{person}{Angela Fan},
  \bibinfo{person}{Christopher Akiki}, \bibinfo{person}{Ellie Pavlick},
  \bibinfo{person}{Suzana Ili{\'c}}, \bibinfo{person}{Daniel Hesslow},
  \bibinfo{person}{Roman Castagn{\'e}}, \bibinfo{person}{Alexandra~Sasha
  Luccioni}, \bibinfo{person}{Fran{\c{c}}ois Yvon}, {et~al\mbox{.}}}
  \bibinfo{year}{2022}\natexlab{}.
\newblock \showarticletitle{Bloom: A 176b-parameter open-access multilingual
  language model}.
\newblock \bibinfo{journal}{\emph{arXiv preprint arXiv:2211.05100}}
  (\bibinfo{year}{2022}).
\newblock


\bibitem[Yong et~al\mbox{.}(2022)]%
        {yong2022bloom}
\bibfield{author}{\bibinfo{person}{Zheng-Xin Yong}, \bibinfo{person}{Hailey
  Schoelkopf}, \bibinfo{person}{Niklas Muennighoff},
  \bibinfo{person}{Alham~Fikri Aji}, \bibinfo{person}{David~Ifeoluwa Adelani},
  \bibinfo{person}{Khalid Almubarak}, \bibinfo{person}{M~Saiful Bari},
  \bibinfo{person}{Lintang Sutawika}, \bibinfo{person}{Jungo Kasai},
  \bibinfo{person}{Ahmed Baruwa}, {et~al\mbox{.}}}
  \bibinfo{year}{2022}\natexlab{}.
\newblock \showarticletitle{Bloom+ 1: Adding language support to bloom for
  zero-shot prompting}.
\newblock \bibinfo{journal}{\emph{arXiv preprint arXiv:2212.09535}}
  (\bibinfo{year}{2022}).
\newblock


\bibitem[Zhao et~al\mbox{.}(2023)]%
        {zhao2023survey}
\bibfield{author}{\bibinfo{person}{Wayne~Xin Zhao}, \bibinfo{person}{Kun Zhou},
  \bibinfo{person}{Junyi Li}, \bibinfo{person}{Tianyi Tang},
  \bibinfo{person}{Xiaolei Wang}, \bibinfo{person}{Yupeng Hou},
  \bibinfo{person}{Yingqian Min}, \bibinfo{person}{Beichen Zhang},
  \bibinfo{person}{Junjie Zhang}, \bibinfo{person}{Zican Dong},
  {et~al\mbox{.}}} \bibinfo{year}{2023}\natexlab{}.
\newblock \showarticletitle{A survey of large language models}.
\newblock \bibinfo{journal}{\emph{arXiv preprint arXiv:2303.18223}}
  (\bibinfo{year}{2023}).
\newblock


\end{thebibliography}



\end{document}